\PassOptionsToPackage{dvipsnames,table}{xcolor}
\documentclass[authoryear]{dreamxreport}

\usepackage[utf8]{inputenc}
\usepackage{url}
\usepackage{nicefrac}
\usepackage{multirow}
\usepackage{makecell}
\usepackage{placeins}

\definecolor{dreamxprimary}{HTML}{0096FA}
\definecolor{dreamxtext}{HTML}{17242D}
\definecolor{dreamxbackground}{HTML}{F4F9FD}
\definecolor{dreamxheadergray}{HTML}{677991}
\definecolor{DeepNavy}{HTML}{243B53}
\definecolor{GroupRow}{HTML}{F8FAFC}
\definecolor{MutedText}{HTML}{94A3B8}
\hypersetup{linkcolor=dreamxprimary,citecolor=dreamxprimary,urlcolor=dreamxprimary}

\newif\ifshowrevisions
\showrevisionsfalse
\DeclareRobustCommand{\rev}[1]{%
  \ifshowrevisions
    \leavevmode{\color{blue}\hypersetup{linkcolor=blue,citecolor=blue,urlcolor=blue}#1}%
  \else
    #1%
  \fi
}
\newcommand{\revisioncolor}{%
  \ifshowrevisions
    \color{blue}\hypersetup{linkcolor=blue,citecolor=blue,urlcolor=blue}%
  \fi
}

\newcommand{\method}{Urbex}
\newcommand{\bev}{BEV}
\newcommand{\iou}{IoU}
\newcommand{\resulttablesize}{\small}
\newcommand{\cmark}{\ensuremath{\checkmark}}
\newcommand{\na}{\textcolor{MutedText}{\scriptsize n/a}}
\newcommand{\tablegroup}[1]{\textcolor{DeepNavy}{\textit{#1}}}
\newcommand{\missingfigurebox}[1]{%
  \fbox{%
    \begin{minipage}[c][0.28\linewidth][c]{0.9\linewidth}
      \centering
      \small Figure file not found: \texttt{\detokenize{#1}}
    \end{minipage}%
  }%
}

\DreamXLogo{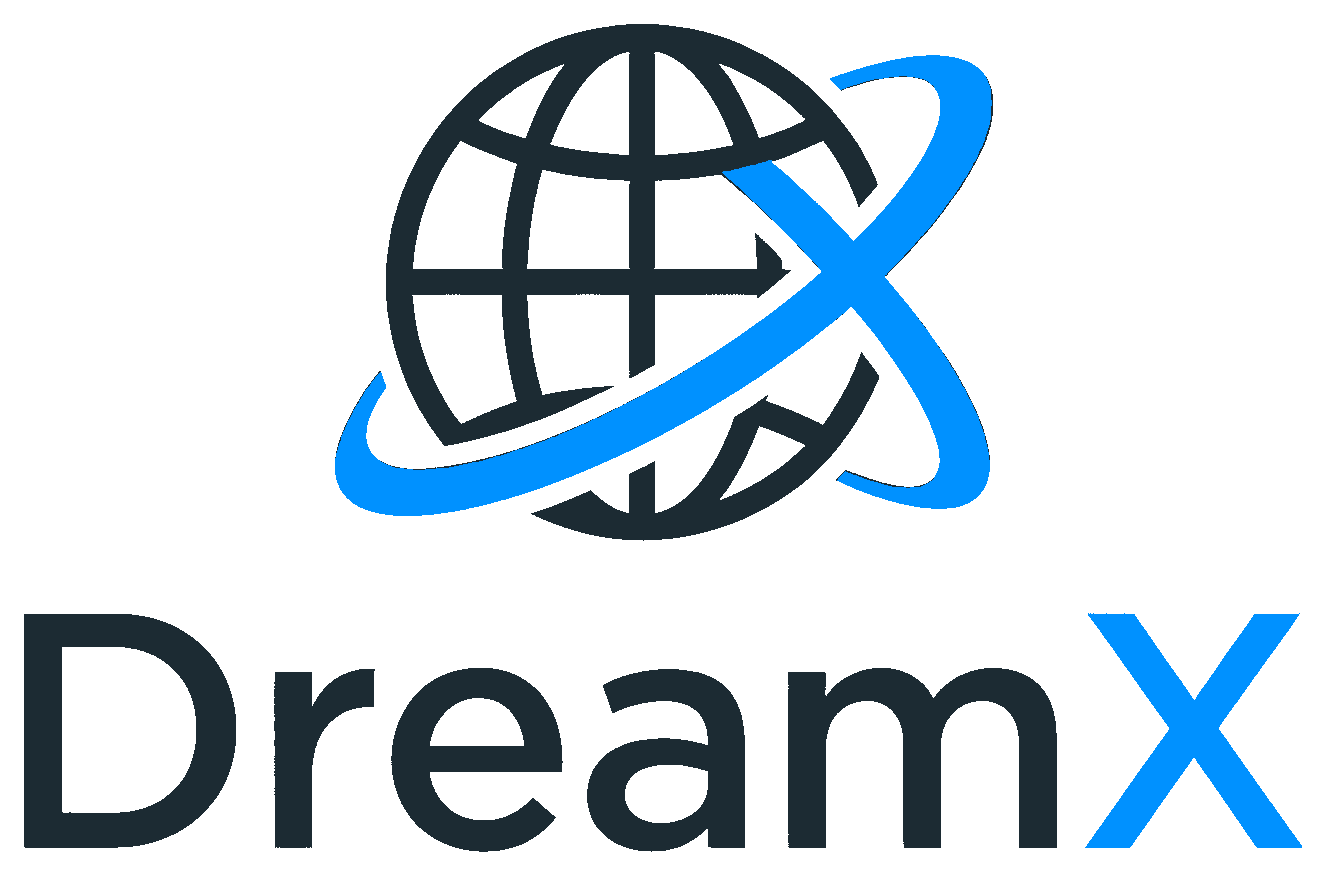}
\makeatletter
\renewcommand{\dreamx@logonode}{%
  \IfFileExists{\dreamx@logo}{%
    \node[anchor=north east, inner sep=0pt, at=(frame.north east), xshift=-13.2pt, yshift=-13.2pt]
      {\includegraphics[height=35.7pt]{\dreamx@logo}};%
    \path[line width=0pt] ([yshift=0.55cm-0.2pt]frame.north);%
  }{%
    \ClassWarning{dreamxreport}{Logo `\dreamx@logo' not found; continuing without it}%
  }%
}
\makeatother

\DreamXRunningTitle{Urbex}
\DreamXHeaderLeft{DreamX Team}
\DreamXHeaderRight{\today}
\DreamXPDFAuthor{Yu Wang, Rui Dai, Wei Chen, Yi Wang, Bo Dang, Kaikui Liu, Xiangxiang Chu, Yansheng Li}
\title{Urbex: Agentic Spatial Grounding for City-Scale 3D Scenes}
\DreamXAuthorPlacement{title}
\author[1,2]{Yu Wang}
\author[2,\dagger]{Rui Dai}
\author[1]{Wei Chen}
\author[2]{Yi Wang}
\author[1]{Bo Dang}
\DreamXAuthorLineBreak
\author[2]{Kaikui Liu}
\author[2,\ddagger]{Xiangxiang Chu}
\author[1,\ddagger]{Yansheng Li}
\affiliation[1]{Wuhan University}
\affiliation[2]{AMAP, Alibaba Group}
\contribution[\dagger]{Project leader}
\contribution[\ddagger]{Corresponding authors}

\reportabstract{
City-scale 3D grounding is commonly formulated as candidate ranking, which requires precomputed instance proposals and becomes inefficient in large outdoor scenes. We propose \method, an agentic vision-language framework that instead casts grounding as active spatial search. \method{} represents a city-scale 3D scene as an interactive multi-view environment, allowing a VLM agent to query landmarks, zoom into local regions, render oblique views, and commit a grounded bounding box. We optimize this tool-use policy with reinforcement learning on CityRefer, using smooth localization rewards and lightweight shaping rewards for efficient exploration. Experiments show that \method{} achieves the best Hit@1 and IoU on CityRefer, supports candidate-free grounding without ground-truth boxes, and is substantially faster than the strongest city-scale ranking baseline. Zero-shot results on STPLS3D-Refer further indicate improved cross-scene robustness.
}
\metadata[Code]{\url{https://github.com/AMAP-ML/Urbex}}
\date{\today}

\begin{document}

\maketitle

\section{Introduction}

Grounding objects in 3D scenes from natural language descriptions \citep{liu2025survey} is a core capability for embodied intelligence, with applications in autonomous driving \citep{cheng2023language}, robotic navigation \citep{roh2022languagerefer}, and aerial inspection \citep{li20253eed}. While indoor 3D grounding has been extensively studied \citep{chen2020scanrefer,achlioptas2020referit3d}, city-scale 3D grounding is still underexplored \citep{miyanishi2023cityrefer}. Outdoor urban scenes are much larger than indoor scans, often containing hundreds of thousands to tens of millions of points per scene \citep{hu2022sensaturban}. The associated referring expressions are also more abstract, as a query may describe a small object through a named building, a road direction, or a distant landmark, e.g., ``a white car parked next to the orange car to the left of the West Midlands Police Custody Suite.'' Solving such queries requires both fine-grained visual recognition and reasoning over long-range spatial cues.

\begin{figure}[t]
  \centering
  \IfFileExists{figures/intro.pdf}{%
    \includegraphics[width=1\linewidth]{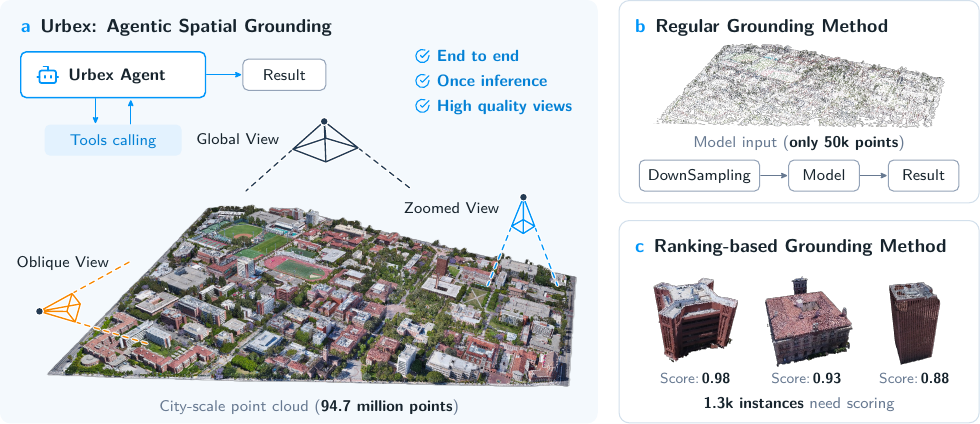}%
  }{%
    \missingfigurebox{figures/intro.pdf}%
  }
  \caption{Overview of \method{}. Unlike regular grounding methods that downsample city-scale point clouds or ranking-based methods that score many precomputed instances, \method{} formulates grounding as agentic spatial search. A VLM agent actively interacts with the scene through tool calls, moving between global, zoomed, and oblique views before committing the final target location.}
  \label{fig:overview}
\end{figure}

Existing 3D grounding methods designed for indoor scenes \citep{chen2020scanrefer,wu2023eda} typically downsample raw point clouds to a fixed size, such as 50,000 points. 
When applied to city-scale scenes, this strategy discards substantial fine-grained geometry and often leads to suboptimal performance, as illustrated in Figure~\ref{fig:overview}b. 
To avoid processing the entire point cloud, recent city-scale methods instead formulate grounding as candidate ranking \citep{miyanishi2023cityrefer,li2025cityanchor}, as shown in Figure~\ref{fig:overview}c. 
They first obtain object proposals from ground-truth annotations or an off-the-shelf instance segmentation pipeline \citep{jiang2020pointgroup}, and then compute text-instance matching scores for each candidate. 
Although effective in reducing input scale, this formulation remains tightly coupled to precomputed candidates: its accuracy depends on proposal quality, while its inference cost grows with the number of candidates to be scored \citep{luo20223d}. 
More importantly, it reduces grounding to a static retrieval problem, whereas localization in a vast urban environment is naturally sequential and exploratory \citep{chaplot2020object,chen2019touchdown}.

In this work, we instead formulate city-scale grounding as an active spatial search problem. Given a referring expression and an initial global observation, a model should parse the target attributes and spatial cues, choose where to inspect next, acquire new evidence, and decide when the evidence is sufficient to commit an answer. Recent tool-using language and vision-language agents \citep{bai2025qwen3,achiam2023gpt} provide a natural substrate for this formulation, as they can reason over visual observations and interact with external tools \citep{yao2023react,schick2023toolformer,ji2026thinking}. However, generic tool use is not sufficient. City-scale 3D grounding requires a task-specific interface that exposes the scene at the right level of abstraction, allowing the agent to move between global context and local detail without exhaustively enumerating all 3D instances.

We propose \method, an agentic spatial grounding framework for city-scale 3D scenes. As illustrated in Figure~\ref{fig:overview}a, \method{} starts from a global bird's-eye view (BEV), uses landmark queries to establish coarse spatial anchors, zooms into promising regions for fine-grained inspection, renders oblique views for 3D confirmation, and finally commits a box that is projected back to world coordinates. \method{} represents each scene as an interactive multi-view environment built around a BEV image and a consistent view-to-world coordinate mapping. The VLM reasons over the accumulated interaction history and either calls another tool or commits a final bounding box in a known \bev{}-aligned view. The system then projects the box back to world coordinates and evaluates it against the target instance. We train \method{} with reinforcement learning on CityRefer \citep{miyanishi2023cityrefer}. The grounding episode is modeled as a multi-step decision process. A terminal reward encourages high overlap with the target and provides a dense center-distance signal when overlap is initially sparse. In addition, tool-level shaping rewards guide intermediate behavior, such as zooming toward the target region. This objective optimizes both final localization accuracy and the agent's intermediate search policy. Our contributions are summarized as follows:
\begin{itemize}
    \item We propose \method{}, an agentic framework that reformulates city-scale 3D grounding from exhaustive candidate ranking to active spatial search.
    \item We design a city-aware interactive tool interface that enables a VLM agent to search large urban scenes through landmark querying, local zooming, and oblique-view verification, and optimize this multi-step policy with reinforcement learning.
    \item \method{} achieves the best Hit@1 and IoU on CityRefer, substantially improves online inference efficiency over the strongest ranking baseline, and demonstrates strong zero-shot robustness on STPLS3D-Refer.
\end{itemize}

\section{Related work}

\subsection{3D Scene Grounding}

3D scene grounding aims to localize a specific object within a 3D environment based on a natural-language referring expression. A large body of prior work has predominantly focused on indoor scenes, where early methods typically encode entire point clouds or multi-view images, fuse them with a language encoder, and rank candidate instances via cross-modal matching \citep{chen2020scanrefer,achlioptas2020referit3d,huang2022multi}. Subsequent research has significantly advanced this paradigm by incorporating relation modeling, multi-view context, and transformer-based architectures \citep{yuan2021instancerefer,wu2023eda}, and exploring proposal-free, bottom-up localization to avoid error accumulation from segmentation models \citep{luo20223d,chen2022language}. Recent zero-shot and open-world methods further exploit foundation models, scene parsing, multi-view reasoning, or scene graphs to reduce dependence on task-specific supervised training \citep{huang2025openground,liu2025viewongraph,zhang2026uniground,gu2024conceptgraphs}. These methods are highly relevant to the goal of reducing proposal dependence, but they are primarily designed for indoor or object-centric 3D scenes and do not directly address city-scale landmark reasoning, very large search spaces, or online efficiency over dense urban instances.

City-scale 3D grounding presents challenges that are substantially different from those in indoor scenes. Outdoor urban environments cover much larger spatial extents, contain many repetitive instances such as vehicles and buildings, and often require reasoning over long-range spatial cues and named landmarks. As a result, existing city-scale methods typically adopt a static candidate-ranking pipeline, where instance boxes are obtained in advance, multimodal features are extracted offline, and a scoring model ranks all candidates during inference \citep{miyanishi2023cityrefer,li2025cityanchor}. Although this formulation avoids directly processing the entire city-scale point cloud, it remains constrained by the quality of precomputed proposals, requires costly feature preprocessing, and becomes inefficient when many candidates must be scored. More importantly, it treats localization as a one-shot retrieval problem. In contrast, \method{} frames city-scale grounding as sequential spatial interaction, actively searching the scene through a small number of targeted observations.

\subsection{Agentic Vision-Language Models}

Modern vision-language models (VLMs) have demonstrated exceptional capabilities in visual question answering, fine-grained reasoning, and instruction following \citep{zheng2026code2world,liu2023visual,chu2024mobilevlm,bai2025qwen3,achiam2023gpt}. RL further improves reasoning capabilities \citep{chu2026gpg}. Building upon these foundation models, recent research has evolved from passive perception to active, agentic systems capable of calling tools, manipulating graphical interfaces, and interpreting complex visual environments \citep{suris2023vipergpt,gupta2023visual,xie2024osworld,ji2026thinking}. Instead of outputting a single-pass prediction, such agentic frameworks interact with their surroundings through a continuous loop of observation, reasoning, action, and feedback \citep{yao2023react,schick2023toolformer,ma2026skillclaw,yuan2026videostar,wang2026urban,wang2026looparena}.

However, much of the agentic VLM literature focuses either on 2D tasks, such as web navigation and open-ended image reasoning, or on indoor embodied navigation \citep{ma2023sqa3d,huang2024an}. Recent work has begun to formulate 3D grounding itself as an agentic or sequential process, including dynamic 2D-to-3D reconstruction with visual tools \citep{wang2026tab}, active cognition over open-world object lookup structures \citep{huang2025openground}, graph-guided VLM reasoning over 3D scene structure \citep{liu2025viewongraph}, and task-oriented sequential grounding in indoor scenes \citep{zhang2024sg3d}. While conceptually close to our active-search view, these works operate in indoor RGB-D or graph-structured settings and are not optimized for city-scale 3D environments. \method{} specializes the agentic VLM paradigm for this challenging domain by providing city-aware spatial tools and an RL objective tailored for efficient exploration.

\section{Method}

\method{} formulates city-scale 3D grounding as active spatial search rather than static candidate retrieval. The central design choice is to expose a large 3D scene through a small number of coordinate-aligned visual observations that a VLM can inspect and act upon. We first describe how the environment converts city-scale geometry into a VLM-operable multi-view space, then define the tool-use reasoning loop over these views, and finally present the reinforcement learning objective that turns prompted tool use into a learned search policy.

\begin{figure}[t]
  \centering
  \IfFileExists{figures/method.pdf}{%
    \includegraphics[width=1\linewidth]{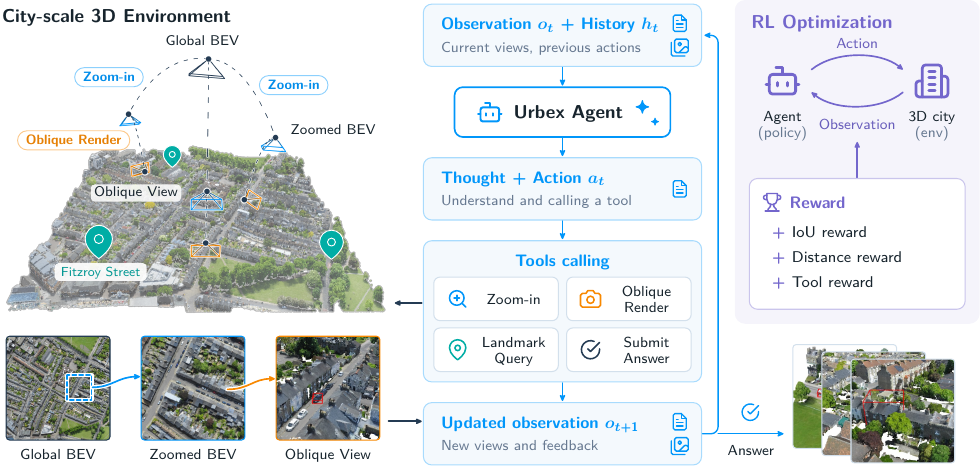}%
  }{%
    \missingfigurebox{figures/method.pdf}%
  }
  \caption{Architecture of \method{}: (left) a multi-view environment exposing global BEV, zoomed, and oblique views; (middle) the policy's reasoning trace and tool calls over observation $o_t$ and interaction history $h_t$; (right) RL under a terminal reward (\iou{} and center-distance) plus tool-level shaping.}
  \label{fig:method}
\end{figure}

\subsection{Interactive environment for city-scale 3D grounding}

Given a city-scale 3D scene $\mathcal{S}$ and a natural-language query $q$, the goal is to localize the referred object $o^\star$ and output its spatial extent. A conventional candidate-ranking method assumes a candidate set $\mathcal{C}=\{o_i\}_{i=1}^{N}$ and learns a score $s(q,o_i)$ for each candidate, making the quality and size of $\mathcal{C}$ central to both accuracy and efficiency \citep{li2025cityanchor,miyanishi2023cityrefer}. \rev{In contrast, \method{} receives no candidate set and exposes the scene through views $\mathcal{V}$ that the agent can extend through tool calls. It predicts a free-form box $\hat b_v$ in a known \bev{}-aligned view $v$, which the environment maps to a world-space extent. Ground-truth boxes provide training rewards and evaluation targets, but are not inference inputs.}

\paragraph{View representation.}
The initial view is a global bird's-eye-view image $I_{\mathrm{global}}$. Each \rev{\bev{}-aligned view $v \in \mathcal{V}$} contains an image observation $I_v$ and a mapping $T_v$ from the view coordinate frame to world coordinates. The agent expresses boxes in a normalized 2D coordinate frame. The environment maintains the corresponding view-to-world transform, so a box predicted in \rev{any such view} can be projected back to world coordinates. This design gives the VLM a simple and consistent spatial interface while preserving metric grounding in the original 3D scene.

\paragraph{City-aware tools.}
\rev{The environment provides three complementary tools for global anchoring, local detail recovery, and 3D verification, keeping the action space compact.}
\textit{Landmark Query} resolves a named landmark from the query, such as a building or road name, and returns its world position and its projection in a selected view. This tool provides global spatial anchors when the expression contains proper nouns. This design is motivated by the geography-aware nature of city-scale referring expressions, where named buildings, roads, and other landmarks provide strong global anchors \citep{miyanishi2023cityrefer,li2025cityanchor,OpenStreetMap}.
\rev{\textit{Zoom-in} maps a selected box to the original high-resolution \bev{} and returns a local crop with a new view identifier and transform. The original BEV has a long side typically exceeding 10,000 pixels, so cropping recovers genuine detail absent from the downsampled global observation, where small objects occupy only a few pixels.}
\textit{Oblique Rendering} renders a selected candidate region from a specified cardinal direction, producing a 3D-aware view that helps verify object shape, height, and spatial placement. The oblique image is used as supporting evidence rather than as the coordinate frame for the final submission. Detailed tool definitions and argument schemas are provided in Table~\ref{tab:appendix_tools}.

\subsection{Agentic reasoning with multi-view tool use}

The policy $\pi_\theta$ is a VLM that receives the current observation $o_t$ and the interaction history $h_t$. The history contains the original query, previous tool calls, returned view identifiers, coordinate conventions, and visual observations. At step $t$, the policy produces a reasoning trace and an action:
\begin{equation}
  (\mathrm{thought}_t, a_t) \sim \pi_\theta(o_t, h_t).
\end{equation}
The action $a_t$ is either a tool call with structured arguments or a final answer. A tool call may specify landmark names, a source view, a 2D crop box, a candidate box, or an oblique rendering direction. The environment executes the tool and appends a textual response and, when applicable, a new image observation to the history.

The resulting inference procedure naturally decomposes into global search, local observation, and candidate confirmation. The agent first parses the expression to identify target attributes and spatial anchors in the \bev{}, then zooms into promising regions to resolve small objects or dense local context, and finally renders oblique views when top-down evidence is insufficient to verify the 3D object. Because every tool response is appended to the interaction history, the model can refer back to previous views and maintain coordinate consistency while switching between global and local evidence. This protocol can be used directly with a prompted VLM, yielding a zero-shot tool-use baseline, but the search space remains large. The model must learn not only how to interpret each view, but also when to query, zoom, render, or stop, which motivates the RL training described next.

\subsection{Reinforcement learning for active spatial search}

We model an episode as a finite-horizon Markov decision process \citep{sutton2018reinforcement}. The state $s_t$ consists of the dialogue history, the current set of views, and tool feedback. The action space includes valid tool invocations and final answer submission. Given an action and a scene, the environment transition is deterministic; trajectory-level stochasticity comes from policy sampling. The episode terminates when the agent submits an answer or reaches the maximum number of assistant turns.

\paragraph{Terminal reward.}
Let $\hat{b}$ be the submitted world-space box and $b^\star$ be the ground-truth target box. We use a smooth terminal reward:
\begin{equation}
  R_{\mathrm{term}}
  =
  \lambda_{\mathrm{iou}} \, \mathrm{IoU}(\hat{b}, b^\star)
  +
  \lambda_{\mathrm{dist}} \exp\!\left(-\frac{d(\hat{b}, b^\star)}{\tau}\right)
  -
  \lambda_{\mathrm{step}} \frac{T}{T_{\max}},
  \label{eq:terminal_reward}
\end{equation}
where $d(\hat{b},b^\star)$ is the center distance in the ground plane, $T$ is the number of assistant turns, and $T_{\max}$ is the interaction budget. The \iou{} term rewards precise localization. The distance term provides a continuous learning signal even when overlap is zero, which is common early in training for vast city-scale scenes and has been shown to be useful in spatial navigation tasks \citep{savva2019habitat}. The step cost discourages unnecessarily long trajectories. We do not use a discrete success bonus; the reward grows smoothly with localization quality rather than introducing a threshold cliff.

\paragraph{Tool-level shaping.}
In addition to the terminal reward, we employ lightweight process rewards \citep{lightman2023let} to shape the intermediate tool-calling behavior \citep{ng1999policy}:
\begin{equation}
  R(\tau) = R_{\mathrm{term}} + \sum_{t=1}^{T} R_{\mathrm{tool}}(s_t,a_t).
\end{equation}
For \textit{Zoom-in}, the reward is positive when the zoom center moves closer to the target compared with the previous zoom and negative when it moves away, with a small cap to prevent it from dominating the terminal objective. For \textit{Oblique Rendering}, the reward is proportional to the overlap between the rendered candidate region and the target region, again capped to keep it as a shaping signal. Invalid tool names, malformed arguments, impossible boxes, or tool execution errors receive negative rewards. These rewards are privileged training signals derived from ground-truth annotations, analogous to reward functions used in simulated navigation or RLVR settings. They are used only for policy optimization and are never included in the agent's text observation, so they do not leak target information at inference time.

\paragraph{Optimization.}
We train \method{} on CityRefer \citep{miyanishi2023cityrefer} entirely via reinforcement learning using Group Relative Policy Optimization \citep{shao2024deepseekmath}. To circumvent the prohibitive memory overhead of a separate value network (critic), it estimates the baseline directly from multiple concurrent rollouts. Specifically, for each input query and initial environment state, the behavior policy $\pi_{\theta_{\mathrm{old}}}$ samples a group of $G$ trajectories $\{\tau_1, \tau_2, \dots, \tau_G\}$. After evaluating the cumulative reward $R(\tau_i)$ for each trajectory, we compute the group-relative advantage:
\begin{equation}
  A_i =
  \frac{R(\tau_i)-\mu_R}{\sigma_R+\epsilon},
  \label{eq:grpo_advantage}
\end{equation}
where $\mu_R$ and $\sigma_R$ are the mean and standard deviation of the group's rewards. The policy is then optimized by maximizing the clipped GRPO objective:
\begin{equation}
\begin{aligned}
  \mathcal{J}_{\mathrm{GRPO}}(\theta)
  &=
  \frac{1}{G}\sum_{i=1}^{G}\frac{1}{T_i}\sum_{t=1}^{T_i}
  \left[
    \min\!\left(
      r_{i,t}(\theta) A_i,\,
      \mathrm{clip}\!\left(r_{i,t}(\theta),1-\epsilon_c,1+\epsilon_c\right) A_i
    \right)
    - \beta D_{\mathrm{KL}}^{i,t}
  \right].
  \label{eq:grpo_objective}
\end{aligned}
\end{equation}
where $r_{i,t}(\theta)=\pi_\theta(a_{i,t}\mid s_{i,t})/\pi_{\theta_{\mathrm{old}}}(a_{i,t}\mid s_{i,t})$, $\epsilon_c$ is the clipping threshold, and $\beta$ controls the token-level KL penalty to the reference policy $\pi_{\mathrm{ref}}$. By relying solely on this RL objective without supervised trajectory demonstrations, the model explores the multi-view environment and learns a task-specific tool-use policy, including when to query landmarks, how large a zoom region should be, whether an oblique view justifies its step cost, and when the accumulated evidence is sufficient to commit a final bounding box.

\section{Experiments}

\subsection{Experimental setup}

We evaluate \method{} along three axes, namely in-domain performance, cross-scene generalization, and component sensitivity. CityRefer \citep{miyanishi2023cityrefer} is used as the main benchmark for comparing grounding accuracy and inference efficiency, while STPLS3D-Refer, our evaluation set constructed from STPLS3D \citep{chen2022stpls3d}, tests whether the learned policy transfers to scenes with different geometry and visual appearance. We convert both datasets into the same \bev{}-based interactive environment and use the same action protocol and evaluation code for all tool-using models. Detailed dataset statistics are provided in Appendix~\ref{sec:datasets}, while training and inference protocols, hyperparameters, and other implementation details are provided in Appendix~\ref{sec:implementation}.

\paragraph{Baselines.}
We compare three groups of methods that stress different aspects of the problem. General-purpose multimodal models, including GPT-5.4 and Qwen3.6-plus, provide strong visual reasoning priors. We connect them to the same \method{} tool environment to test whether a powerful off-the-shelf VLM can solve the task through prompting alone, without task-specific policy optimization. Indoor 3D grounding methods reveal two different failure modes when moved to city-scale scenes. EDA follows the common indoor practice of processing a downsampled point cloud, which discards much of the fine-grained geometry needed for small urban objects \citep{wu2023eda}. InstanceRefer instead operates in a candidate-ranking setting with provided instances, which is easier than direct full-scene localization and therefore performs more reasonably than dense point-cloud grounding methods \citep{yuan2021instancerefer}. City-scale methods such as CityRefer and CityAnchor are closer to our setting, but they still primarily rely on ranking precomputed candidates rather than actively searching the scene \citep{miyanishi2023cityrefer,li2025cityanchor}. Details on each baseline implementation are provided in Appendix~\ref{sec:baselines}.

\begin{figure}[t]
  \centering
  \IfFileExists{figures/cityrefer_qualitative.pdf}{%
    \includegraphics[width=1\linewidth]{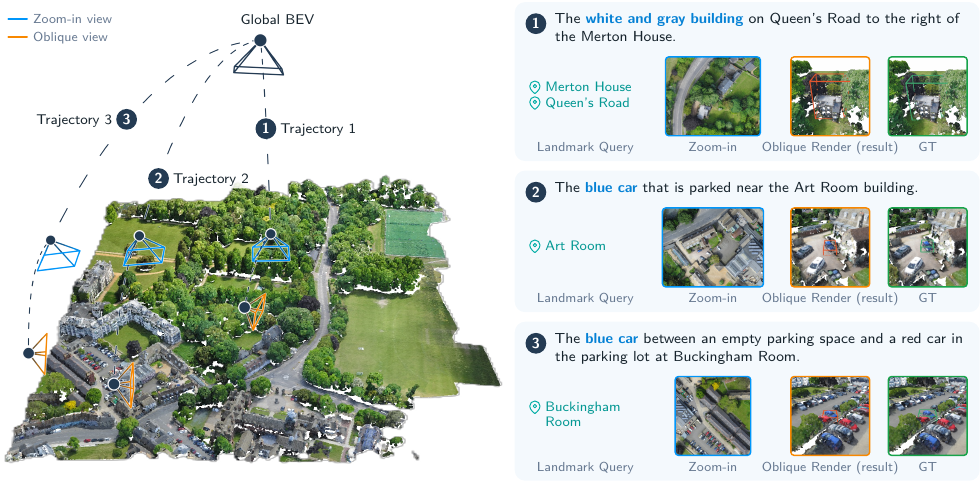}%
  }{%
    \missingfigurebox{figures/cityrefer_qualitative.pdf}%
  }
  \caption{Qualitative examples on CityRefer. The visualization shows how \method{} moves from global \bev{} search to local zoom-in and final grounding in city-scale scenes with dense repeated objects and landmark-based references.}
  \label{fig:cityrefer_qualitative}
\end{figure}

\paragraph{Evaluation settings and metrics.}
We report results on the novel description (ND) and novel object (NO) splits: ND uses unseen descriptions in scenes observed during training, while NO uses targets from held-out scenes that do not appear in training. In \textit{With GT}, candidate-ranking baselines may use the ground-truth candidate instance set and are evaluated with Hit@1, Hit@3, and Hit@5. \method{} does not use candidate boxes during inference even in this setting; it always outputs a free-form \bev{} box, \rev{which is matched to candidates only after prediction: candidates are ranked by 2D \iou{} with the predicted box, and Hit@$k$ records whether the target is among the top $k$, without a minimum overlap requirement.} \rev{This measures target identification; thresholded Hit@1 and direct overlap metrics additionally assess box quality (Appendix~\ref{sec:matching}).} In \textit{Without GT}, methods cannot access ground-truth candidate boxes and must directly output a spatial prediction; we therefore report 2D \iou{}, 3D \iou{}, and point-level F1 against the target. This setting is closer to deployment, where complete and accurate instance boundaries are rarely available. Rows whose metrics would all be inapplicable are omitted from the corresponding subtable; otherwise inapplicable cells are marked as \na{}. GPT-5.4 and Qwen3.6-plus do not expose comparable timing measurements, while InstanceRefer, CityRefer, and CityAnchor require candidate instances; \rev{we separately test CityAnchor with predicted proposals in Appendix~\ref{sec:proposals}.} \rev{Accuracy metrics are percentages.} We report \rev{inference time per example} on a single GPU, including model generation and online tool execution, while excluding offline scene preparation and cached rendering construction for all methods.

\subsection{Main results on CityRefer}

\begin{table}[t]
  \caption{\rev{Main results on CityRefer. \method{} supports candidate-free grounding and efficient inference. Five-run Hit@1 statistics are reported in Table~\ref{tab:repeat_runs}.}}
  \label{tab:cityrefer}
  \centering
  \resulttablesize
  \setlength{\tabcolsep}{4pt}
  \begin{tabular*}{\linewidth}{@{\extracolsep{\fill}}lccccccc@{}}
    \toprule
    \multirow{2}{*}{Method}
    & \multicolumn{3}{c}{ND With GT}
    & \multicolumn{3}{c}{NO With GT}
    & \multirow{2}{*}{Time (s)} \\
    \cmidrule(lr){2-4}\cmidrule(lr){5-7}
    & Hit@1 & Hit@3 & Hit@5
    & Hit@1 & Hit@3 & Hit@5
    & \\
    \midrule
    \rowcolor{GroupRow}
    \multicolumn{8}{@{}l}{\tablegroup{Closed-source VLMs w/ \method{} tools}} \\
    GPT-5.4 & 9.3 & 12.4 & 15.7 & 13.1 & 18.6 & 20.1 & \na \\
    Qwen3.6-plus & 6.4 & 9.6 & 11.9 & 11.5 & 16.2 & 16.8 & \na \\
    \midrule
    \rowcolor{GroupRow}
    \multicolumn{8}{@{}l}{\tablegroup{Indoor 3D grounding methods}} \\
    InstanceRefer~\citep{yuan2021instancerefer} & 1.6 & 5.0 & 10.5 & 4.9 & 15.9 & 24.5 & \textbf{0.6} \\
    EDA~\citep{wu2023eda} & 0.1 & 0.1 & 0.2 & 0.2 & 0.3 & 0.4 & 3.0 \\
    \midrule
    \rowcolor{GroupRow}
    \multicolumn{8}{@{}l}{\tablegroup{Ranking-based city-scale methods}} \\
    CityRefer~\citep{miyanishi2023cityrefer} & 1.6 & 3.9 & 4.8 & 0.6 & 2.1 & 3.4 & \underline{0.7} \\
    CityAnchor~\citep{li2025cityanchor} & \underline{23.4} & \underline{35.3} & \textbf{45.4} & \underline{14.2} & \underline{22.4} & \underline{31.0} & 32.1 \\
    \midrule
    \method{} & \textbf{25.8} & \textbf{39.2} & \underline{42.3} & \textbf{23.9} & \textbf{36.9} & \textbf{40.7} & 3.2 \\
    \bottomrule
  \end{tabular*}

  \vspace{0.6em}
  \begin{tabular*}{\linewidth}{@{\extracolsep{\fill}}lcccccc@{}}
    \toprule
    \multirow{2}{*}{Method}
    & \multicolumn{3}{c}{ND Without GT}
    & \multicolumn{3}{c}{NO Without GT} \\
    \cmidrule(lr){2-4}\cmidrule(lr){5-7}
    & 2D \iou{} & 3D \iou{} & Point F1
    & 2D \iou{} & 3D \iou{} & Point F1 \\
    \midrule
    \rowcolor{GroupRow}
    \multicolumn{7}{@{}l}{\tablegroup{Closed-source VLMs w/ \method{} tools}} \\
    GPT-5.4 & \underline{4.4} & \underline{2.2} & 9.4 & \underline{4.6} & 1.9 & 10.4 \\
    Qwen3.6-plus & 3.0 & 1.8 & \underline{9.5} & 4.3 & \underline{2.1} & \underline{12.1} \\
    \midrule
    \rowcolor{GroupRow}
    \multicolumn{7}{@{}l}{\tablegroup{Indoor 3D grounding methods}} \\
    EDA~\citep{wu2023eda} & 0.2 & 0.2 & 0.3 & 0.5 & 0.4 & 0.8 \\
    \midrule
    \method{} & \textbf{15.6} & \textbf{11.0} & \textbf{20.1} & \textbf{14.6} & \textbf{7.7} & \textbf{17.8} \\
    \bottomrule
  \end{tabular*}
  \vspace{-10px}
\end{table}

Table~\ref{tab:cityrefer} reports the main CityRefer results. GPT-5.4 and Qwen3.6-plus obtain non-trivial accuracy after being connected to the \method{} tool environment, showing that strong VLMs can exploit the interactive views and tool feedback to perform partial active search. However, their performance remains far below \method{}, \rev{showing that these prompted tool-use baselines do not recover the accuracy of the trained policy.} The indoor baselines illustrate why directly transferring indoor 3D grounding pipelines is also insufficient. EDA almost fails on both splits, consistent with the severe information loss caused by downsampling city-scale point clouds before grounding. InstanceRefer performs better because it is evaluated as a simpler candidate-ranking problem with provided instances, but it still remains far below city-scale methods and cannot directly solve candidate-free localization.

Among city-scale baselines, CityRefer is limited when evaluated as a full ranking problem because its original setting assumes that the target appears in a small 10-instance candidate pool rather than requiring discrimination over all scene instances \citep{miyanishi2023cityrefer}. CityAnchor is much stronger because it scores candidates with multimodal large language models, but this design requires applying the scoring model to many instances and therefore incurs high inference cost. In the With-GT setting, \method{} achieves the best Hit@1 and Hit@3 on both test splits and the best Hit@5 on NO. \rev{CityAnchor retains the highest Hit@5 on ND: exhaustive ranking can retain several plausible GT candidates, including a target that \method{} misses during regional search or confuses with a local distractor. Thus, broader candidate recall and precise single-box search have different failure modes (Appendix~\ref{sec:failure_analysis}).}

The advantage becomes clearer in the Without-GT setting, \rev{where candidate-ranking methods require a proposal-generation stage.} General multimodal models can use the same prompted tool interface, but their 2D and 3D \iou{} remain substantially lower than \method{}. \rev{To test a complete proposal-based pipeline, we train SoftGroup~\citep{vu2022softgroup} on 15 non-CityRefer SensatUrban scenes and supply its proposals to CityAnchor. On NO, this pipeline reaches 4.2/2.4/4.7 in BEV \iou{}/3D \iou{}/point F1, versus 14.6/7.7/17.8 for \method{} (Appendix~\ref{sec:proposals}), showing the practical gap for this proposal-based pipeline.}

\paragraph{\rev{Inference efficiency.}}
\method{} is also substantially more efficient than the strongest ranking baseline under the online timing protocol above. CityAnchor, the second-best-performing method in the With-GT comparison, requires about $32.1$ seconds per example because it scores many candidate instances, whereas \method{} takes about $3.2$ seconds. This amounts to an order-of-magnitude inference-speed improvement, since \method{} inspects a small number of informative regions rather than exhaustively scoring all instances. \rev{The tools account for about 0.36 s/query. Building the optional oblique cache takes approximately 2.3 hours across 12 CityRefer scenes; rendering on demand instead gives an estimated 4.4 s/query ($7.3\times$ faster). These comparisons exclude offline scene preparation; Appendix~\ref{sec:latency} separates the measured cache cost from other, unmeasured preparation components.}

\begin{figure}[t]
  \centering
  \IfFileExists{figures/stpls3d_qualitative.pdf}{%
    \includegraphics[width=1\linewidth]{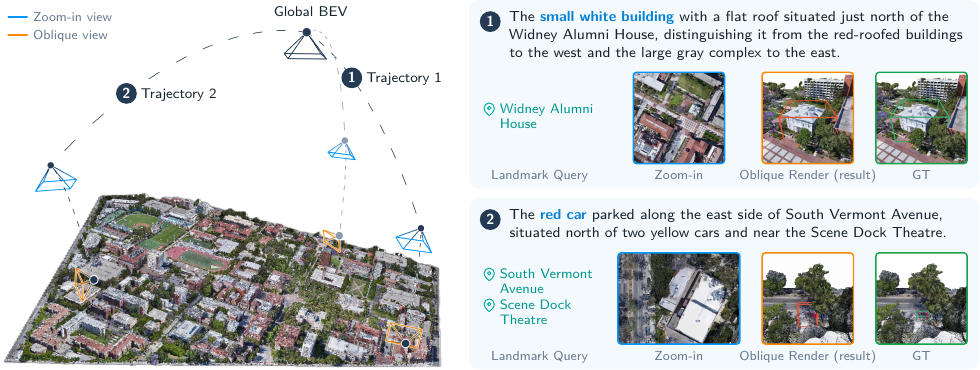}%
  }{%
    \missingfigurebox{figures/stpls3d_qualitative.pdf}%
  }
  \caption{Qualitative examples on STPLS3D-Refer. The visualization highlights zero-shot transfer cases where the same active-search policy anchors, zooms, verifies, and grounds targets in scenes with different geometry and object distributions from CityRefer.}
  \label{fig:stpls3d_qualitative}

\end{figure}

\begin{table}[t]
  \caption{\rev{Zero-shot generalization on STPLS3D-Refer. Table~\ref{tab:repeat_runs} reports five-run Hit@1 statistics.}}
  \label{tab:stpls3d}
  \centering
  \resulttablesize
  \setlength{\tabcolsep}{5pt}
  \begin{tabular*}{\linewidth}{@{\extracolsep{\fill}}lccccccc@{}}
    \toprule
    \multirow{2}{*}{Method}
    & \multicolumn{3}{c}{With GT}
    & \multicolumn{3}{c}{Without GT}
    & \multirow{2}{*}{Time (s)} \\
    \cmidrule(lr){2-4}\cmidrule(lr){5-7}
    & Hit@1 & Hit@3 & Hit@5 & 2D \iou{} & 3D \iou{} & Point F1 & \\
    \midrule
    \rowcolor{GroupRow}
    \multicolumn{8}{@{}l}{\tablegroup{Closed-source VLMs w/ \method{} tools}} \\
    GPT-5.4 & \underline{16.3} & \underline{17.8} & \underline{19.1} & 8.0 & 5.3 & 12.1 & \na \\
    Qwen3.6-plus & 12.0 & 13.8 & 14.3 & \underline{8.1} & \underline{6.4} & \underline{12.2} & \na \\
    \midrule
    \rowcolor{GroupRow}
    \multicolumn{8}{@{}l}{\tablegroup{Indoor 3D grounding methods}} \\
    InstanceRefer~\citep{yuan2021instancerefer} & 3.8 & 7.8 & 12.3 & \na & \na & \na & \textbf{0.6} \\
    EDA~\citep{wu2023eda} & 0.0 & 0.0 & 0.5 & 0.4 & 0.2 & 0.4 & 2.9 \\
    \midrule
    \rowcolor{GroupRow}
    \multicolumn{8}{@{}l}{\tablegroup{Candidate-dependent city-scale methods}} \\
    CityRefer~\citep{miyanishi2023cityrefer} & 1.4 & 2.5 & 3.8 & \na & \na & \na & \underline{1.4} \\
    CityAnchor~\citep{li2025cityanchor} & 2.1 & 4.2 & 6.0 & \na & \na & \na & 47.2 \\
    \midrule
    \method{} & \textbf{27.2} & \textbf{32.1} & \textbf{33.4} & \textbf{13.0} & \textbf{10.0} & \textbf{16.2} & 4.8 \\
    \bottomrule
  \end{tabular*}
  \vspace{-15px}
\end{table}

\subsection{Zero-shot generalization on STPLS3D-Refer}

We next evaluate zero-shot transfer to STPLS3D-Refer. The agent is trained only on CityRefer and is directly applied to STPLS3D-Refer after converting scenes into the same \bev{}-based interactive environment. This setting tests whether the model has learned a reusable spatial search strategy rather than a dataset-specific ranking function. As shown in Table~\ref{tab:stpls3d}, the new scenes are challenging for all compared methods: indoor grounding methods are sensitive to domain shift, and city-scale ranking models degrade when candidate distributions and visual statistics differ from CityRefer. GPT-5.4 and Qwen3.6-plus perform reasonably well after being connected to the \method{} tool environment, again showing that strong VLM reasoning can exploit interactive observations even without task-specific training. However, \method{} achieves the best score on every reported accuracy metric. These aggregate results suggest, but do not by themselves prove, that the RL-trained policy learns reusable search behaviors such as anchoring, zooming, verifying, and stopping. \rev{To examine transfer beyond named geographic anchors, we additionally evaluate 30 manually verified landmark-free re-annotations of STPLS3D-Refer. Without further training, \method{} achieves 23.3 Hit@1 on these expressions, providing complementary evidence that the learned policy transfers to queries expressed through visual attributes and local spatial relations. Appendix~\ref{sec:landmark_robustness} details this evaluation and the landmark-robustness experiments; Figure~\ref{fig:stpls3d_qualitative} illustrates the agent's spatial search and grounding trajectories in the new scenes.}

\subsection{Ablation studies}

Tables~\ref{tab:tool_ablation} and~\ref{tab:reward_ablation} ablate the tool interface and the reward design. Tool variants add \textit{Landmark Query} and \textit{Oblique Rendering} on top of \textit{Zoom-in}, while reward variants retain terminal \iou{} and progressively add distance and tool-level shaping terms.

\begin{table}[!htbp]
  \centering
  \scriptsize
  \begin{minipage}[c]{0.52\linewidth}
    \centering
    \caption{Tool and budget ablation for \method.}
    \label{tab:tool_ablation}
    \setlength{\tabcolsep}{2.5pt}
    \begin{tabular*}{\linewidth}{@{\extracolsep{\fill}}cccccc@{}}
      \toprule
      Zoom & Landmark & Oblique & Turns & Hit@3 & 2D \iou{} \\
      \midrule
      -- & -- & -- & 0 & 10.5 & 2.1 \\
      \cmark & -- & -- & 8 & 15.5 & 4.1 \\
      \cmark & \cmark & -- & 8 & 28.2 & 8.9 \\
      \cmark & \cmark & \cmark & 7 & 34.8 & 11.0 \\
      \cmark & \cmark & \cmark & 8 & \textbf{36.9} & \textbf{14.6} \\
      \cmark & \cmark & \cmark & 9 & 36.8 & 13.2 \\
      \bottomrule
    \end{tabular*}
  \end{minipage}
  \hfill
  \begin{minipage}[c]{0.44\linewidth}
    \centering
    \caption{Reward ablation for \method.}
    \label{tab:reward_ablation}
    \setlength{\tabcolsep}{2.5pt}
    \begin{tabular*}{\linewidth}{@{\extracolsep{\fill}}lcc@{}}
      \toprule
      Reward setting & Hit@3 & 2D \iou{} \\
      \midrule
      \iou{} only & 22.8 & 6.7 \\
      \iou{} + Distance & 30.6 & 10.9 \\
      \iou{} + Tool-level & 28.4 & 9.6 \\
      Full reward & \textbf{36.9} & \textbf{14.6} \\
      \bottomrule
    \end{tabular*}
  \end{minipage}
\end{table}

The ablation results show that each component contributes a distinct part of the active-search behavior. \rev{Without tools, the agent sees the entire downsampled global map but reaches only 10.5 Hit@3 and 2.1 2D \iou{}, compared with 36.9 and 14.6 using all tools.} Adding \textit{Zoom-in} improves both metrics, confirming that high-resolution local evidence is important for small and densely arranged urban objects. The largest tool gain comes from adding \textit{Landmark Query}: Hit@3 rises from $15.5$ to $28.2$, and 2D \iou{} more than doubles from $4.1$ to $8.9$. This supports our premise that named buildings, roads, and other geographic anchors are central to city-scale referring expressions. In contrast to indoor scenes where a local crop may already contain most relevant context, city-scale queries often require first locating the correct neighborhood before fine-grained recognition becomes meaningful.

\textit{Oblique Rendering} further improves the result by helping the agent verify candidate regions with 3D-aware visual evidence. With all tools enabled, increasing the budget from 7 to 8 turns improves Hit@3 from $34.8$ to $36.9$ and 2D \iou{} from $11.0$ to $14.6$, suggesting that one additional interaction is often useful for confirmation before submission. Increasing the budget to 9 turns does not improve Hit@3 and slightly lowers 2D \iou{}, indicating that excessive interaction can introduce unnecessary exploration or late-stage box adjustments. We therefore use the 8-turn setting as the main configuration, which provides the best accuracy-efficiency trade-off in this ablation. This also suggests that \method{} benefits from a bounded search process rather than unbounded deliberation.

The reward ablation confirms the same pattern. Terminal \iou{} alone reaches $22.8$ Hit@3 and $6.7$ 2D \iou{}; adding distance shaping improves them to $30.6$ and $10.9$, tool-level shaping is also beneficial, and the full reward performs best. Distance shaping is especially useful when early predictions have no overlap with the target, while tool-level rewards make tool use more stable. Their combination shows that localization quality, distance progress, and efficient tool use are complementary.

\subsection{\rev{Analysis}}

\paragraph{\rev{Tool-use policy.}}
\rev{We further examine the learned policy using a fixed Landmark Query $\rightarrow$ Zoom-in $\rightarrow$ Oblique Rendering $\rightarrow$ Answer workflow (Appendix~\ref{sec:scripted}). In the reported experiments, the base Qwen model reaches only 2.7 Hit@1, whereas RL-trained \method{} reaches 10.0 under the same tool order. The fixed workflow still requires the model to predict tool arguments and the final box. The improvement therefore suggests that training helps the model use the observed scene to make spatial decisions, such as selecting a crop and locating the target within it. The workflow is also useful for a stronger prompted VLM: GPT-5.4 maintains similar Hit@1 while reducing token use from 16,321 to 7,734. This suggests that the learned tool sequence provides an efficient way to organize search, while the trained model contributes to how the tools are used.}

\paragraph{\rev{Spatial search.}}
\rev{The landmark experiments further show the importance of geographic cues for city-scale search. On CityRefer-NO expressions with named landmarks, the 10 m coordinate-noise setting lowers Hit@1 by only 1.2 points, whereas missing or incorrect landmarks cause drops exceeding ten points (Appendix~\ref{sec:landmark_robustness}). This suggests that a moderately inaccurate anchor can still guide the search, but losing the reference required by the expression is substantially more harmful. Figure~\ref{fig:failure_cases} illustrates errors at different stages of this process. In (a), the first zoom excludes the target, and the agent selects another car within the crop. In (b), the zoom contains the target, but the agent selects vehicles in a different parking row. Both predictions remain unchanged after oblique inspection. The first case calls for revisiting the search region, while the second requires closer comparison of nearby instances using the query's spatial relations. These examples motivate improving both region selection and local verification (Appendix~\ref{sec:failure_analysis}).}

\paragraph{\rev{Localization quality.}}
\rev{We also examine the quality of the free-form boxes produced by \method{} (Appendix~\ref{sec:matching}). On CityRefer-NO, replacing the original IoU matching rule with positive-overlap filtering or CIoU changes Hit@1 by at most 0.29 points, indicating that target identification is stable under these matching rules. However, additionally requiring the first-ranked target to have BEV \iou{} $\geq0.25$ reduces Hit@1 from 23.53 to 18.71. A prediction can therefore identify the correct candidate while still including substantial surrounding context. The area-ratio statistics show the same issue: the predicted-to-target area ratio has median 1.40 and P90 9.92, indicating that some boxes are considerably larger than the target. These results suggest that refining the box after target selection could further improve localization. We report direct \iou{} and point F1 alongside Hit@$k$ to assess both target identification and the accuracy of the predicted spatial extent.}

\section{Conclusion}

We presented \method{}, an agentic framework for city-scale 3D grounding. 
Rather than ranking a large set of precomputed candidates, \method{} formulates grounding as active spatial search in an interactive multi-view environment. We further optimize this tool-use policy with reinforcement learning under a limited interaction budget, encouraging accurate and efficient exploration. Experiments on CityRefer show that \method{} achieves strong grounding accuracy, supports candidate-free localization, and substantially improves online inference efficiency over ranking-based baselines. Zero-shot results on STPLS3D-Refer further suggest that active spatial search provides a robust alternative to static candidate retrieval for large-scale 3D scene understanding.

\bibliographystyle{dreamxplainnat}
\bibliography{references}

\clearpage
\reportappendix
\begingroup
\let\FloatBarrier\relax
\patchcmd{\section}{-3.5ex}{-2.75ex}{}{}
\patchcmd{\section}{2.3ex}{1.75ex}{}{}
\patchcmd{\subsection}{-3.25ex}{-2.5ex}{}{}
\patchcmd{\subsection}{1.5ex}{1.0ex}{}{}
\patchcmd{\paragraph}{3.25ex}{2.5ex}{}{}
\setkeys{Gin}{height=0.88\textheight,keepaspectratio}
\appendix

\begingroup
\raggedbottom
\makeatletter
\setlength{\@fptop}{0pt}
\makeatother
\setlength{\parskip}{4pt}
\setlength{\textfloatsep}{12pt plus 2pt minus 2pt}
\setlength{\floatsep}{10pt plus 2pt minus 2pt}
\setlength{\intextsep}{10pt plus 2pt minus 2pt}

\section{Implementation Details}
\label{sec:implementation}

\subsection{Hyperparameters and Hardware}
\label{sec:hyperparams}

\method{} is initialized from Qwen3-VL-4B-Instruct \citep{bai2025qwen3}. Training and evaluation use the same
city-tool interface, action schema, coordinate convention, and execution code for both CityRefer and STPLS3D-Refer.
Unless otherwise specified, each trajectory allows at most eight assistant turns: up to seven tool calls and one final
answer turn.

\paragraph{Training configuration.}
\method{} is trained on the CityRefer training split with GRPO \citep{shao2024deepseekmath}: rollout batch size
$256$, group size $8$, $300$ optimization steps, and one PPO epoch per rollout batch. The actor learning rate is
$1\times10^{-6}$, the KL coefficient is $0.001$, and bf16 training freezes the vision encoder. The reward combines
final \iou{}, center-distance shaping, a step cost, and tool-level shaping rewards, with
$\lambda_{\mathrm{iou}}=1.0$, $\lambda_{\mathrm{dist}}=0.3$, $\tau=80$m, and $\lambda_{\mathrm{step}}=0.05$ in
Equation~\ref{eq:terminal_reward}.

\paragraph{Hardware and timing protocol.}
Training uses $16$ NVIDIA H20 GPUs. \rev{Local-model inference and timing use a single NVIDIA H20 GPU; closed-source VLMs are accessed through their APIs.} We report
average online inference time per query, including model generation and online tool execution, while excluding
offline scene preprocessing such as cached renderings and candidate-pool preparation for all methods.

\subsection{Tool Interface}

Table~\ref{tab:appendix_tools} summarizes the tool interface shared by prompted VLM baselines and \method{}. Before
execution, the environment validates tool names, argument types, box bounds, and view identifiers.

\begin{table}[htbp]
  \caption{Definitions, inputs, and outputs of the city-aware tools.}
  \label{tab:appendix_tools}
  \centering
  \footnotesize
  \setlength{\tabcolsep}{3pt}
  \renewcommand{\arraystretch}{1.15}
  \begin{tabularx}{\linewidth}{@{}>{\raggedright\arraybackslash}p{0.17\linewidth}
      >{\raggedright\arraybackslash}p{0.27\linewidth}
      >{\raggedright\arraybackslash}p{0.28\linewidth}
      >{\raggedright\arraybackslash}X@{}}
    \toprule
    Tool & Definition & Inputs & Outputs \\
    \midrule
    \textit{Landmark Query}
    & Resolves a named urban landmark, such as a building, road, or facility, and uses it as a global spatial anchor.
    & \texttt{landmark\_name}: text span or proper noun; \texttt{source\_view\_id}: view in which the returned
      projection should be expressed; optional \texttt{landmark\_type} when the name is ambiguous.
    & Match status; matched landmark name and type; world-space center or footprint when available; projection of the
      landmark into the requested view. \\
    \midrule
    \textit{Zoom-in}
    & \rev{Maps a region from an existing \bev{} view to the original high-resolution image and crops it, exposing detail absent from the global observation.}
    & \texttt{source\_view\_id}: global or local \bev{} view; \texttt{crop\_box}: normalized 2D box; optional
      \texttt{scale} or requested output resolution.
    & Local \bev{} image; new \texttt{view\_id}; updated view-to-world transform; crop extent in the source and world
      coordinate frames. \\
    \midrule
    \textit{Oblique Rendering}
    & Renders an agent-selected candidate region from a cardinal direction to provide 3D-aware verification of shape,
      height, facade appearance, and spatial placement.
    & \texttt{source\_view\_id}: \bev{} view containing the candidate; \texttt{candidate\_box}: normalized 2D region;
      \texttt{direction}: one of north, south, east, or west; optional \texttt{margin} around the region.
    & Oblique image observation; rendering metadata including direction and covered world region; supporting evidence
      for verification. The oblique view is not used as the coordinate frame for final box submission. \\
    \bottomrule
  \end{tabularx}
\end{table}

\subsection{Baseline Settings}
\label{sec:baselines}

Closed-source VLMs use the same \method{} tool environment and maximum-turn budget but no task-specific policy
optimization. GPT-5.4 and Qwen3.6-plus receive the same action format,
coordinate convention, tool descriptions, final-answer format, and tool-output history, then submit a final
\bev{}-aligned box. These baselines test whether strong off-the-shelf VLMs can solve city-scale grounding with
prompting and tool access alone.

InstanceRefer \citep{yuan2021instancerefer} and EDA \citep{wu2023eda} are adapted from indoor 3D grounding pipelines
by following their released preprocessing and downsampling each city-scale point cloud to $40{,}000$ points.
InstanceRefer is evaluated with provided instances in its native candidate-ranking setting and is therefore omitted
from candidate-free Without-GT evaluation; EDA uses the same point budget. This preserves the indoor pipelines while
exposing the loss of fine-grained geometry needed for small or repetitive urban objects.

CityRefer and CityAnchor \citep{miyanishi2023cityrefer,li2025cityanchor} are retrained on our CityRefer split with
the original hyperparameters, optimizer, schedule, and loss weights unless noted. For CityAnchor, we use the released
ROI-localization and candidate-scoring pipeline, lower the ROI threshold to $0.1$ to retain candidates under our
larger pool, use landmark features from the retrained CityRefer language module as officially recommended, and include
online candidate scoring in its timing.

\begin{figure}[htbp]
  \centering
  \IfFileExists{figures/prompt.pdf}{%
    \includegraphics[width=0.85\linewidth]{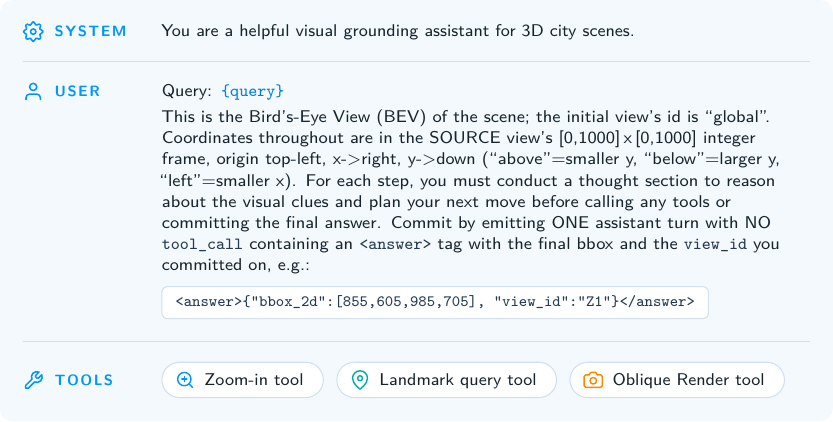}%
  }{%
    \missingfigurebox{figures/prompt.pdf}%
  }
  \caption{Prompt template for tool-using VLM agents, specifying the task, tools, coordinates, action format, and
  final box format shared by prompted VLM baselines and \method{}.}
  \label{fig:appendix_prompt}
\end{figure}

\section{Dataset Details}
\label{sec:datasets}

\subsection{CityRefer Data Preparation}
\label{sec:cityrefer_data}

CityRefer \citep{miyanishi2023cityrefer}, built on SensatUrban \citep{hu2022sensaturban}, is our in-domain benchmark.
We remove duplicate \texttt{(scene, query)} rows and low-quality annotations with visibly incorrect instance labels or
boxes, then create train, validation, ND, and NO splits. ND contains unseen descriptions in training scenes; NO tests
held-out-scene generalization. Seven scenes are reserved for NO, and the remaining five are split across train,
validation, and ND with a fixed random seed, yielding $2{,}744$ training, $549$ validation, $1{,}098$ ND, and
$1{,}396$ NO queries.

\begin{table}[htbp]
  \revisioncolor
  \centering\small
  \caption{Scale of the 12 CityRefer scenes used in our experiments. Test queries combine ND and NO.}
  \label{tab:cityrefer_scale}
  \begin{tabular}{lrr}
    \toprule
    Statistic & Total across 12 scenes & Per-scene range \\
    \midrule
    Area (km$^2$) & 1.594 & 0.029--0.165 \\
    Points (millions) & 877.4 & 12.1--116.3 \\
    GT candidate objects & 2,234 & 31--367 \\
    OSM landmarks & 299 & 9--51 \\
    Test queries & 2,494 & 14--556 \\
    \bottomrule
  \end{tabular}
\end{table}

\rev{Small objects are difficult to distinguish in a downsampled global \bev{}. We therefore retain a high-resolution \bev{} for each scene and use \textit{Zoom-in} to retrieve detailed local crops during search.}

\subsection{STPLS3D-Refer Construction}
\label{sec:stpls3d_data}

STPLS3D-Refer is a cross-domain zero-shot benchmark for testing whether a CityRefer-trained policy transfers to new
urban geometry, appearance, and geography. It is derived from public STPLS3D RealWorld photogrammetry point clouds
\citep{chen2022stpls3d} and used only for evaluation. Because STPLS3D lacks CityRefer-style expressions and object
instances, we combine geometric instance extraction, geographic registration, OpenStreetMap landmarks
\citep{OpenStreetMap}, and VLM-assisted expression generation to build a diagnostic, semi-automatic benchmark.

\paragraph{Source scenes and target categories.}
From four STPLS3D drone-photogrammetry scenes, we retain the two with reliable public geographic context: University
of Southern California Park Campus (USC, $94.7$M points, $1083{\times}807$m) and Wrigley Marine Science Center (WMSC,
$9.98$M points, $490{\times}293$m). We exclude one scene dominated by a single mega-building and one without
verifiable public geographic context. Targets are limited to \textit{Building} and \textit{Vehicle}.

\paragraph{Instance extraction.}
Since STPLS3D provides semantic labels but not instances, we cluster semantic masks with DBSCAN: buildings use planar
$xy$ coordinates with eps$=2.0$m, min\_samples$=80$, and footprint filter $[4,5000]$m$^2$; vehicles use $xyz$ with
eps$=0.6$m, min\_samples$=30$, and footprint filter $[1.5,30]$m$^2$. From more than $1{,}000$ pseudo-instances, we
remove fragmented clusters, merged objects, ground/noise artifacts, and projection-ambiguous candidates through
geometric filters and manual inspection. The verified pool contains $59$ buildings/$62$ vehicles in USC and $11$
buildings/$17$ vehicles in WMSC, with axis-aligned 3D boxes and point masks computed for each retained cluster.

\paragraph{Geo-registration and landmarks.}
Each scene is registered to WGS-84 by fitting a first-order affine transform from $5$--$7$ ground-control points
selected against recent satellite imagery, with RMS residuals of $0.72$m for USC and $2.19$m for WMSC. We query
OpenStreetMap features using tags including \texttt{building}, \texttt{amenity}, \texttt{highway}, \texttt{leisure},
\texttt{tourism}, \texttt{office}, and \texttt{natural}, project them back to local coordinates, and match landmarks
to building instances when applicable. The two scenes contain $278$ real OSM landmarks.

\paragraph{Scene conversion and baseline compatibility.}
Each scene is converted to the CityRefer interface: global \bev{} image, view-to-world mapping, Zoom-in crops,
optional oblique renderings, and annotations for candidate-based and candidate-free evaluation. We also export
candidate metadata and per-instance features required by CityAnchor-style pipelines \citep{li2025cityanchor}; ranking
methods use explicit candidates, while \method{} receives only the visual tool environment and submits a free-form
\bev{} box.

\paragraph{Referring-expression generation and filtering.}
For each target, we render a centered \bev{} crop together with oblique side views from the four cardinal
directions, mark the target and nearby same-class distractors in all rendered views, and give the target color,
size, distractor relations, and nearest OSM landmarks to Qwen3.6-plus to generate three concise
CityRefer-style expressions. The oblique views supply 3D-aware cues such as building height, facade appearance, and
relative layout that are difficult to disambiguate from the \bev{} alone. Automatic filters check length, meta-language, landmark-name validity,
projection validity, and token-level duplication. The pipeline first yields more than $1{,}000$ preliminary
target-expression samples; manual inspection then removes unclear targets, incorrect landmarks, severe instance
artifacts, and ambiguous descriptions, leaving $445$ high-confidence landmark-grounded samples.

\begin{table}[htbp]
  \centering
  \small
  \setlength{\tabcolsep}{5pt}
  \caption{Statistics of STPLS3D-Refer. ``Inst.'' counts unique candidate instances, and ``Lndmk.'' counts named OSM landmarks.}
  \label{tab:stpls3d_refer_stats}
  \begin{tabular}{lrrrrr}
    \toprule
    Scene & Inst. (Bldg./Veh.) & Building Expr. & Vehicle Expr. & Total Expr. & Lndmk. \\
    \midrule
    USC  & $59 / 62$ & $177$ & $184$ & $361$ & $272$ \\
    WMSC & $11 / 17$ & $33$  & $51$  & $84$  & $6$   \\
    \midrule
    Total & $70 / 79$ & $210$ & $235$ & $445$ & $278$ \\
    \bottomrule
  \end{tabular}
\end{table}

\FloatBarrier
\section{Additional Qualitative Examples}
\label{sec:qualitative}

Figure~\ref{fig:appendix_cityrefer_more} provides additional CityRefer examples, and
Figure~\ref{fig:appendix_agent_reasoning} shows full agent trajectories from query and global \bev{} observation to
Landmark Query, Local Zoom-in, Oblique Rendering, and final box prediction.

\begin{figure}[htbp]
  \centering
  \IfFileExists{figures/cityrefer_more_qualitative.pdf}{%
    \includegraphics[width=1\linewidth]{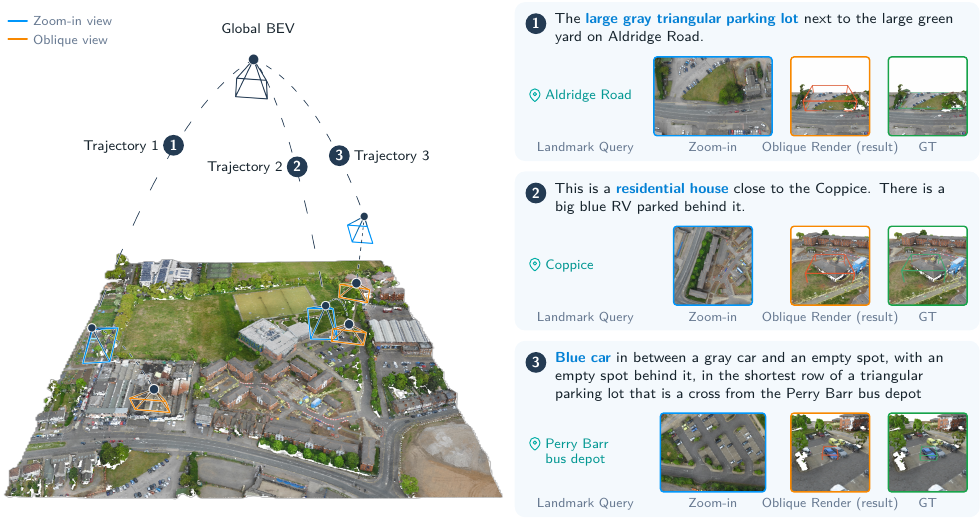}%
  }{%
    \missingfigurebox{figures/cityrefer_more_qualitative.pdf}%
  }
  \caption{Additional qualitative examples on the CityRefer NO split.}
  \label{fig:appendix_cityrefer_more}
\end{figure}

\begin{figure}[htbp]
    \centering
    \includegraphics[width=\textwidth]{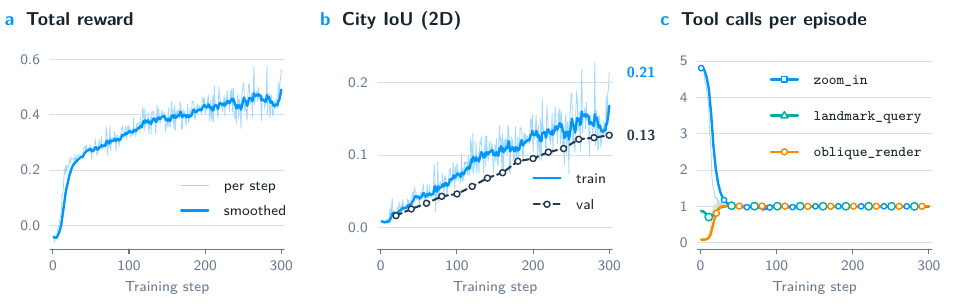}
    \caption{\textbf{Training dynamics on CityRefer.}
    (a)~Total task reward. (b)~Train/validation 2D grounding \iou{} with periodic validation markers and final-value
    annotations. (c)~Per-episode tool-call frequency, showing convergence from over-calling \emph{zoom\_in} and
    under-using \emph{oblique\_render} to one call per tool.}
    \label{fig:appendix_train_dynamics}
\end{figure}

\begin{figure}[htbp]
  \centering
  \IfFileExists{figures/agent_reasoning_process.pdf}{%
    \includegraphics[width=1\linewidth]{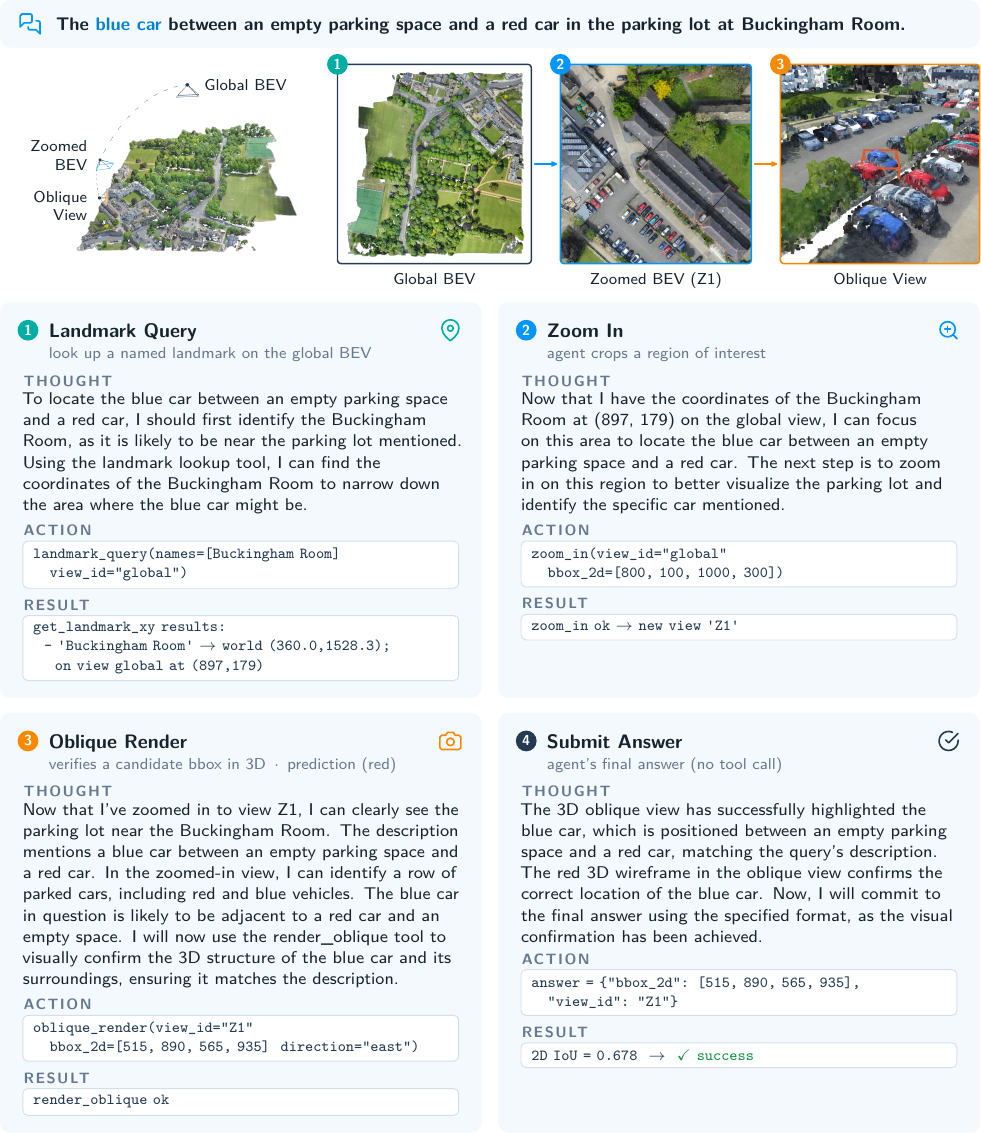}%
  }{%
    \missingfigurebox{figures/agent_reasoning_process.pdf}%
  }
  \caption{Example agent reasoning processes from the initial query and global scene observation to landmark
  anchoring, local zoom-in, oblique verification, and final box commitment.}
  \label{fig:appendix_agent_reasoning}
\end{figure}

Figure~\ref{fig:appendix_train_dynamics} summarizes the first $300$ CityRefer optimization steps. Total reward rises
from slightly negative values to a stable positive regime, while train and validation 2D \iou{} improve together with
a small, stable gap, suggesting generalization rather than memorization. Tool-use statistics show the policy moving
from inefficient exploration to a disciplined sequence: early training overuses \emph{zoom\_in} (peak $\approx4.7$
calls, first-$30$-step mean $2.6$) and underuses \emph{oblique\_render} (mean $\approx0.6$), but later converges to
using each tool roughly once per episode. This matches the Landmark Query~$\rightarrow$~Local Zoom-in~$\rightarrow$~
Oblique Rendering pattern shown in Figure~\ref{fig:appendix_agent_reasoning}.

\FloatBarrier
\begingroup
\revisioncolor
\section{Controlled Experiments and Diagnostic Analysis}
\label{sec:diagnostics}

We report additional experiments on policy execution, landmark robustness, proposal dependence, matching rules,
localization quality, and scene preparation. Unless stated otherwise, accuracy values are percentages.
Tables~\ref{tab:cityrefer} and~\ref{tab:stpls3d} retain the original evaluation run; repeated-run summaries below are
reported separately rather than mixing means with single-run metrics.

\subsection{Repeated Decoding Runs}
\label{sec:repeat_runs}

Table~\ref{tab:repeat_runs} summarizes five stochastic decoding runs of the same trained checkpoint on fixed test
sets. The reported uncertainty is the sample standard deviation across runs, in percentage points. It measures
inference-time sampling variation, not variability across independently trained GRPO policies. An approximate
95\% Student-$t$ interval for the ND mean is
$25.62 \pm t_{4,0.975}(0.33/\sqrt{5}) = [25.21,26.03]$, using $t_{4,0.975}\simeq2.776$.
This interval lies above CityAnchor's reported 23.4, supporting a margin relative to the observed decoding
variation. It is not a paired significance test: it neither captures test-sample uncertainty nor accounts for
training or evaluation variation in the baseline. Independent training seeds and paired per-query comparisons
would be needed to assess those sources of uncertainty.

\begin{table}[htbp]
  \revisioncolor
  \centering\small
  \caption{Hit@1 over five decoding runs of one checkpoint (mean $\pm$ sample standard deviation). Baselines are the strongest comparison results in the original main tables.}
  \label{tab:repeat_runs}
  \begin{tabular}{lrrl}
    \toprule
    Dataset & \method{} Hit@1 & Comparison Hit@1 & Comparison method \\
    \midrule
    CityRefer-ND & $25.62\pm0.33$ & 23.4 & CityAnchor \\
    CityRefer-NO & $23.53\pm0.73$ & 14.2 & CityAnchor \\
    STPLS3D-Refer & $26.34\pm0.58$ & 16.3 & GPT-5.4 \\
    \bottomrule
  \end{tabular}
\end{table}

\subsection{Fixed Tool Workflows and Learned Execution}
\label{sec:scripted}

After RL training, a common trajectory follows Landmark Query $\rightarrow$ Zoom-in $\rightarrow$ Oblique
Rendering $\rightarrow$ Answer (L-Z-O). This order emerges during learning rather than being prescribed in the
training protocol. We extract it and enforce it at inference to form a scripted baseline with the same tool
interface. The model still predicts tool arguments and the final box; ``fixed'' constrains the tool sequence,
not the spatial coordinates. Table~\ref{tab:scripted} compares the reported free and fixed-policy results.

\begin{table}[htbp]
  \revisioncolor
  \centering\small
  \caption{Free and fixed L-Z-O tool policies. The fixed workflow is extracted from the learned policy; tokens are reported per query.}
  \label{tab:scripted}
  \setlength{\tabcolsep}{5pt}
  \begin{tabular}{llrrrr}
    \toprule
    Model & Policy & Hit@1 & Hit@3 & Hit@5 & Tokens/query \\
    \midrule
    Qwen3-VL-4B base & Fixed & 2.7 & 5.0 & 6.0 & 6,576 \\
    \method{} RL & Free & 23.5 & 37.0 & 40.9 & 7,187 \\
    \method{} RL & Fixed & 10.0 & 16.0 & 22.0 & 6,786 \\
    GPT-5.4 & Free & 13.1 & 18.6 & 20.1 & 16,321 \\
    GPT-5.4 & Fixed & 13.5 & 23.0 & 26.5 & 7,734 \\
    \bottomrule
  \end{tabular}
\end{table}

Under the fixed workflow, RL-trained \method{} reaches 10.0 Hit@1 versus 2.7 for the base model, indicating that
the trained model contributes beyond access to a useful tool sequence. Free \method{} reaches 23.5, but this gap
should not be attributed solely to choosing a different order: most free trajectories already follow L-Z-O.
Externally forcing the sequence also changes the interaction context in which arguments and answers are
generated. The comparison shows that replaying a tool order does not recover the full learned execution policy;
it does not isolate tool ordering from argument generation or prompting effects.

For GPT-5.4, the fixed workflow maintains similar Hit@1 (13.5 versus 13.1), improves Hit@3/5 from 18.6/20.1 to
23.0/26.5, and reduces token use from 16,321 to 7,734. Thus, the discovered workflow is also a useful efficiency
prior for an untuned VLM, even though it does not reproduce \method{}'s full accuracy.

\FloatBarrier
\subsection{Landmark Availability and Construction Bias}
\label{sec:landmark_robustness}

We separate CityRefer-NO into 1,302 expressions with named landmarks and 94 without them. For landmark-guided
expressions, \emph{Missing} returns no match, \emph{Noisy 10 m} perturbs the correct coordinate by 10 m, and
\emph{Wrong} returns a same-scene landmark at least 50 m away. Table~\ref{tab:landmark_robustness} reports the
resulting Hit@1. Landmark-free rows use their own expressions and are not perturbations of the same queries.

\begin{table}[htbp]
  \revisioncolor
  \centering\small
  \caption{Landmark robustness (Hit@1, \%). Dashes denote inapplicable landmark perturbations. The 30-example landmark-free STPLS3D set is a separate diagnostic.}
  \label{tab:landmark_robustness}
  \setlength{\tabcolsep}{4pt}
  \begin{tabular}{lrrrrr}
    \toprule
    Model / expression subset & $N$ & Clean & Missing & Noisy 10 m & Wrong \\
    \midrule
    \multicolumn{6}{l}{\textit{CityRefer-NO}} \\
    \method{}, with landmarks & 1,302 & 23.5 & 12.7 & 22.3 & 11.3 \\
    \method{}, without landmarks & 94 & 31.2 & -- & -- & -- \\
    \midrule
    \multicolumn{6}{l}{\textit{STPLS3D-Refer}} \\
    GPT-5.4, with landmarks & 445 & 16.3 & 5.7 & 15.3 & 7.3 \\
    \method{}, with landmarks & 445 & 26.3 & 10.7 & 25.8 & 13.9 \\
    \method{}, landmark-free diagnostic & 30 & 23.3 & -- & -- & -- \\
    \bottomrule
  \end{tabular}
\end{table}

On CityRefer-NO, 10 m coordinate noise changes landmark-guided Hit@1 by 1.2 points, while missing or wrong matches
cause much larger reductions of 10.8 and 12.2 points. Accurate retrieval remains important when a landmark is
needed to distinguish similar targets across a scene. The 31.2 Hit@1 on expressions without named landmarks
shows that some targets can instead be identified through visual attributes and local relations; it does not
imply that removing necessary landmarks from the other expressions is harmless.

All original STPLS3D-Refer expressions are landmark-grounded because both generation and filtering use OSM cues.
To probe this construction bias, we re-annotate 30 targets without named landmarks and manually verify that each
expression uniquely identifies its target among same-class instances. \method{} reaches 23.3 Hit@1 on this
separate diagnostic. On the original 445 expressions, it exceeds GPT-5.4 under each matched perturbation, although
missing and wrong landmarks substantially degrade both models. These results provide limited evidence of
landmark-free transfer; a 30-example targeted set does not establish general robustness on an independently
constructed, landmark-independent benchmark.

\subsection{Grounding with Predicted Instance Proposals}
\label{sec:proposals}

A practical ranking pipeline can replace ground-truth candidates with predicted instances. We train
SoftGroup~\citep{vu2022softgroup} on 15 non-CityRefer SensatUrban scenes and use its proposals as CityAnchor's
candidate pool. The reported CityRefer-NO proposal recall is 35.6\% at \iou{} $\geq0.25$.
Table~\ref{tab:proposal_pipeline} reports the resulting pipeline alongside the GT-candidate reference and
proposal-free \method{}.

\begin{table}[htbp]
  \revisioncolor
  \centering\small
  \caption{CityRefer-NO proposal comparison. All metrics, including overlap and point F1, are percentages. Hit@$k$ uses method-dependent candidate matching; direct IoU/F1 assess localization without a GT candidate pool.}
  \label{tab:proposal_pipeline}
  \setlength{\tabcolsep}{4pt}
  \begin{tabular}{llrrrrrr}
    \toprule
    Method & Proposals & Hit@1 & Hit@3 & Hit@5 & BEV IoU & 3D IoU & Point F1 \\
    \midrule
    CityAnchor & Ground truth & 14.2 & 22.4 & 31.0 & -- & -- & -- \\
    CityAnchor & SoftGroup & 6.5 & 13.0 & 15.0 & 4.2 & 2.4 & 4.7 \\
    \method{} & None & 23.5 & 37.0 & 40.9 & 14.6 & 7.7 & 17.8 \\
    \bottomrule
  \end{tabular}
\end{table}

The proposal pipeline has substantially lower direct localization quality than \method{}: BEV \iou{} is 4.2
versus 14.6, 3D \iou{} is 2.4 versus 7.7, and point F1 is 4.7 versus 17.8. Its Hit@1/3/5 are 6.5/13.0/15.0,
compared with 14.2/22.4/31.0 for CityAnchor supplied with GT candidates. Predicted candidates lack GT instance
identities, so their success requires geometric matching; these Hit@$k$ values should not be conflated with the
GT-identity protocol. The direct IoU/F1 comparison provides complementary evidence that \method{} avoids the
proposal-quality bottleneck. It does not imply that all proposal generators would have the same recall.

\FloatBarrier
\subsection{Candidate Matching and Box-Extent Diagnostics}
\label{sec:matching}

For a free-form prediction $\hat b$ and GT candidate boxes $\{b_i\}_{i=1}^N$, standard With-GT evaluation ranks
candidates in descending order of $\mathrm{IoU}_{\mathrm{BEV}}(\hat b,b_i)$. A hit occurs when the target's identity
is among the top $k$. GT boxes are used only for this post-inference scoring; \method{} neither selects from them
nor receives them as input. Ranking alone imposes no minimum target overlap, so it primarily measures
identification rather than precise extent.

\paragraph{Sensitivity to the ranking rule.}
Table~\ref{tab:matching_sensitivity} re-scores predictions from the five decoding runs on CityRefer-NO using the
original IoU rule, a variant that excludes zero-overlap candidates, and CIoU, which adds center-distance and
aspect-ratio penalties. Hit@1 varies by at most 0.29 points and Hit@3 by at most 0.22. This supports stability of
identification under the tested ranking rules. Positive-overlap filtering removes zero-IoU ties, but it does
not directly diagnose oversized boxes; stable ranking alone cannot establish tight localization.

\begin{table}[htbp]
  \revisioncolor
  \centering\small
  \caption{Matching-rule sensitivity on CityRefer-NO (five-run mean, \%).}
  \label{tab:matching_sensitivity}
  \begin{tabular}{lrrr}
    \toprule
    Matching rule & Hit@1 & Hit@3 & Hit@5 \\
    \midrule
    Original IoU & 23.53 & 36.97 & 40.86 \\
    IoU with positive-overlap filtering & 23.53 & 36.87 & 40.49 \\
    CIoU & 23.24 & 36.75 & 43.18 \\
    \bottomrule
  \end{tabular}
\end{table}

\paragraph{Target-IoU-thresholded Hit@1.}
We additionally require the target to rank first and the free-form box to satisfy
$\mathrm{IoU}_{\mathrm{BEV}}(\hat b,b^\star)\geq\eta$, for $\eta\in\{0.25,0.5\}$:
\begin{equation}
  \mathrm{Hit@1}_{\eta}
  = \frac{100}{Q}\sum_{j=1}^{Q}
  \mathbf{1}\!\left[\operatorname{rank}_j(o_j^\star)=1\right]
  \mathbf{1}\!\left[\mathrm{IoU}_{\mathrm{BEV}}(\hat b_j,b_j^\star)\geq\eta\right].
\end{equation}
Here $Q$ is the number of test queries. Table~\ref{tab:thresholded_hit1} shows that imposing 0.25 target IoU reduces
Hit@1 by 4.82 points on NO and 5.24 points on ND. Loose extents therefore have a measurable effect. These are
thresholded Hit@1 diagnostics, not thresholded Hit@3/5 results.

\begin{table}[htbp]
  \revisioncolor
  \centering\small
  \caption{Target-IoU-thresholded Hit@1 (\%). The target must rank first under the original IoU matching rule.}
  \label{tab:thresholded_hit1}
  \begin{tabular}{lrrr}
    \toprule
    Split & No IoU threshold & Target IoU $\geq0.25$ & Target IoU $\geq0.5$ \\
    \midrule
    CityRefer-NO & 23.53 & 18.71 & 11.46 \\
    CityRefer-ND & 25.62 & 20.38 & 12.61 \\
    \bottomrule
  \end{tabular}
\end{table}

\paragraph{Candidate overlap and area ratio.}
We count candidates overlapped by each free-form prediction, first at any positive IoU and then at IoU
$\geq0.25$ (Table~\ref{tab:overlap_counts}). The mean number of positive overlaps is 6.5 on NO and 4.6 on ND,
but the median number reaching 0.25 IoU is one on both splits. Thus, multiple weak overlaps are common, whereas
a typical prediction does not strongly overlap multiple candidates. The predicted-to-target area ratio
$\mathrm{area}(\hat b)/\mathrm{area}(b^\star)$ has medians of 1.40 and 1.56, with P90 values of 9.92 and 7.21
(Table~\ref{tab:area_ratio}). This reveals a tail of oversized predictions, particularly consequential for
small targets. These diagnostics support separating target identification from box quality; they do not
establish that oversizing is absent.

\begin{table}[htbp]
  \revisioncolor
  \centering
  \begin{minipage}[t]{0.48\linewidth}

  \centering\small
  \setlength{\tabcolsep}{4pt}
  \caption{Number of candidate boxes overlapped per prediction.}
  \label{tab:overlap_counts}
  \begin{tabular}{lrrrr}
    \toprule
    & \multicolumn{2}{c}{IoU $>0$} & \multicolumn{2}{c}{IoU $\geq0.25$} \\
    \cmidrule(lr){2-3}\cmidrule(lr){4-5}
    Split & Mean & Median & Mean & Median \\
    \midrule
    CityRefer-NO & 6.5 & 4 & 0.93 & 1 \\
    CityRefer-ND & 4.6 & 4 & 0.82 & 1 \\
    \bottomrule
  \end{tabular}
  \end{minipage}\hfill
  \begin{minipage}[t]{0.48\linewidth}

  \centering\small
  \caption{Predicted-to-target BEV area ratio (unitless).}
  \label{tab:area_ratio}
  \begin{tabular}{lrrr}
    \toprule
    Split & Median & P75 & P90 \\
    \midrule
    CityRefer-NO & 1.40 & 3.39 & 9.92 \\
    CityRefer-ND & 1.56 & 3.48 & 7.21 \\
    \bottomrule
  \end{tabular}
  \end{minipage}
\end{table}

\FloatBarrier
\subsection{City-Scale Search and Disambiguation Failures}
\label{sec:failure_analysis}

We analyze two factors that complicate city-scale grounding: the distance between the target and its landmark,
and the number of same-class distractors within 50 m. Table~\ref{tab:failure_factors} reports the trajectory
analysis, visualized in Figure~\ref{fig:failure_diagnostics}. Regional-search success measures whether the selected
Zoom-in region reaches the target area before local disambiguation. Local-disambiguation success measures
correct identification among nearby instances after the correct area is reached.

\begin{table}[htbp]
  \revisioncolor
  \centering\small
  \caption{Reported trajectory diagnostics (success, \%). The two panels measure different stages of grounding.}
  \label{tab:failure_factors}
  \begin{tabular}{lrrr}
    \toprule
    Target--landmark distance & ND & NO & STPLS3D \\
    \multicolumn{4}{l}{\textit{Regional-search success}} \\
    \midrule
    $<50$ m & 40.0 & 34.3 & 42.2 \\
    $50$--$100$ m & 31.1 & 32.6 & 26.6 \\
    $100$--$200$ m & 27.3 & 29.1 & 21.7 \\
    $\geq200$ m & 14.3 & 22.5 & 10.0 \\
    \midrule
    Same-class distractors within 50 m & ND & NO & STPLS3D \\
    \multicolumn{4}{l}{\textit{Local-disambiguation success}} \\
    \midrule
    0 & 64.7 & 56.7 & 29.5 \\
    1--3 & 43.1 & 39.1 & 31.2 \\
    4--9 & 41.8 & 36.5 & 2.4 \\
    $\geq10$ & 25.1 & 14.4 & 0.0 \\
    \bottomrule
  \end{tabular}
\end{table}

Regional-search success decreases with landmark distance on all three datasets. A distant landmark gives a
less precise cue when an expression describes qualitative spatial relations rather than a metric displacement.
Local-disambiguation success also falls markedly in dense same-class neighborhoods: on ND it drops from 64.7\%
without nearby same-class distractors to 25.1\% with at least ten. The STPLS3D trend is not monotonic in the first
two bins (29.5\% versus 31.2\%), but the dense bins are substantially harder. These are associations across query
groups; they do not isolate a causal effect of distance or density, and no bin-level uncertainty is reported.

These factors help explain the ND Hit@5 gap. CityAnchor can keep several plausible GT candidates in its top five,
including a target that \method{} misses during regional search. Even after reaching the correct region,
\method{} can confuse nearby repeated instances. This is consistent with search and disambiguation errors rather
than evidence of a fixed visual-recognition ceiling. Adaptive initial crops, distractor-aware verification, and
boundary refinement target different parts of this failure process.

\begin{figure}[htbp]
  \revisioncolor
  \centering
  \includegraphics[width=0.9\linewidth]{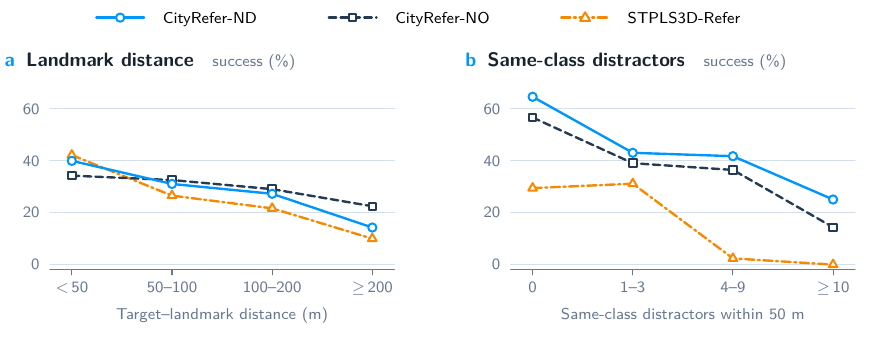}
  \caption{Aggregate trajectory diagnostics from Table~\ref{tab:failure_factors}. Left: success versus target--landmark distance. Right: success versus same-class distractor count within 50 m. These curves visualize the reported bin rates, not individual failure trajectories; per-bin sample counts and error bars are unavailable.}
  \label{fig:failure_diagnostics}
\end{figure}

\FloatBarrier
\begin{samepage}
\paragraph{Representative failure trajectories.}
Figure~\ref{fig:failure_cases} shows two saved CityRefer-NO trajectories: a zoom that excludes the target and a
wrong-row prediction despite a zoom containing the target. In both cases, the agent commits the same box after
one oblique inspection. These selected zero-IoU examples illustrate distinct failure stages; they do not estimate
failure frequencies or establish the causes of the aggregate trends.
\par\end{samepage}

\begin{figure}[!htbp]
  \revisioncolor
  \centering
  \includegraphics[width=0.94\linewidth]{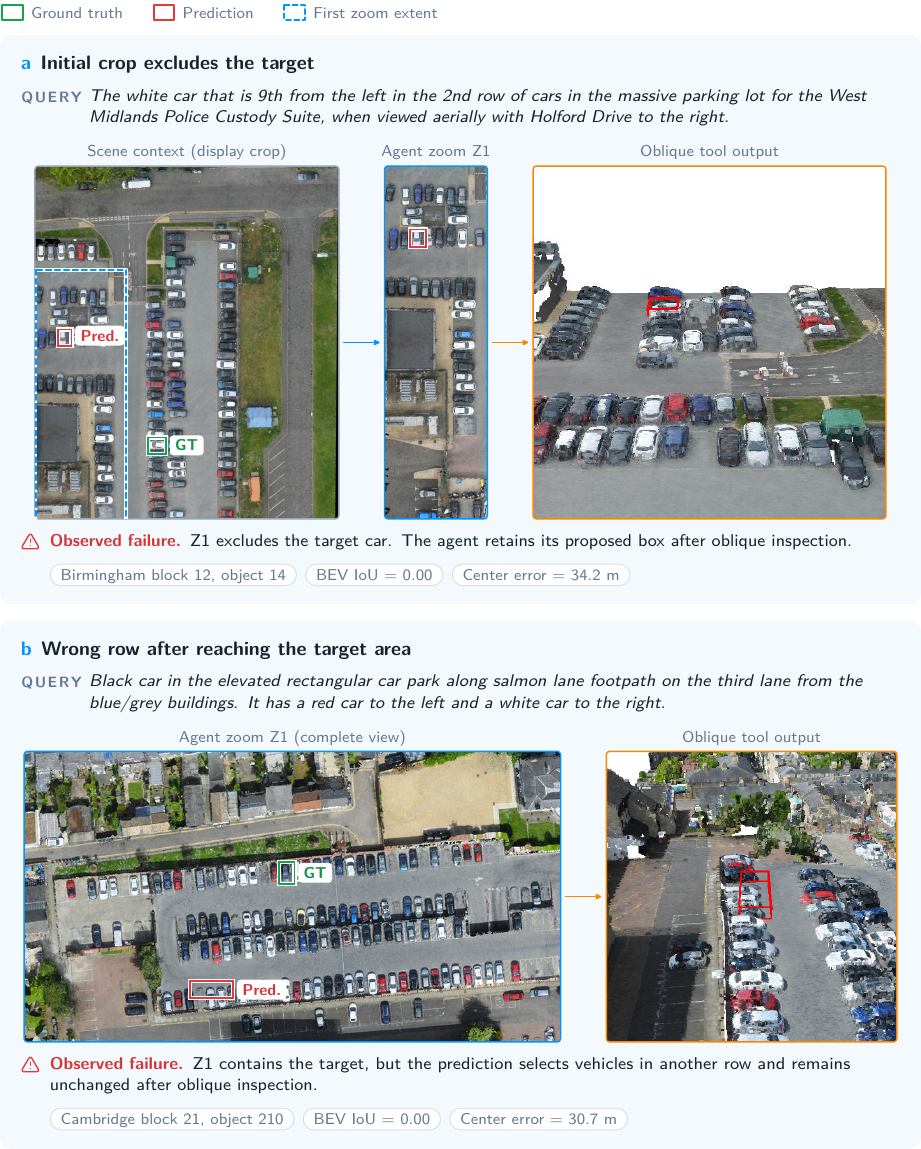}
  \caption{Qualitative failure trajectories on CityRefer-NO. (a) The first zoom excludes the annotated target.
  (b) The zoom includes the target, but the prediction selects vehicles in a different parking row.
  Scene context in (a) is an offline display crop of the high-resolution BEV. Zoom and oblique panels preserve
  the complete recorded tool images; GT and prediction overlays on BEV views are added for diagnosis.
  Green denotes GT, red denotes the prediction, and dashed blue denotes the first zoom extent.
  Both runs retain their proposed box after oblique inspection and obtain zero BEV IoU.}
  \label{fig:failure_cases}
\end{figure}

\FloatBarrier
\subsection{Inference efficiency}
\label{sec:latency}

Table~\ref{tab:latency_breakdown} separates online tool costs, full-query latency, and the one-time oblique-cache
construction cost. Cached Landmark Query, Zoom-in, and Oblique Rendering together take about 0.36 s, included in
\method{}'s 3.2 s/query. CityAnchor's 32.1 s/query also excludes offline preparation and uses pre-extracted
point-cloud and landmark features. The approximately $10\times$ ratio therefore compares warm online inference
under the same accounting scope.

\begin{table}[htbp]
  \revisioncolor
  \centering\small
  \caption{CityRefer latency and optional scene-cache preparation. The on-demand total is an estimate; cache construction covers all 12 scenes.}
  \label{tab:latency_breakdown}
  \begin{tabular}{lr}
    \toprule
    Component & Latency / preparation time \\
    \midrule
    Landmark Query & $<1$ ms \\
    Zoom-in from high-resolution BEV & 70 ms \\
    Oblique Rendering from cache & 292 ms \\
    All three tools & $\approx0.36$ s \\
    \method{}, full model and cached tools & 3.2 s/query \\
    CityAnchor, warm online inference & 32.1 s/query \\
    \midrule
    Optional oblique cache, all 12 scenes & $\approx2.3$ hours (one time) \\
    Additional latency with on-demand oblique views & $\approx1.2$ s/query \\
    \method{}, estimated on-demand total & $\approx4.4$ s/query \\
    \bottomrule
  \end{tabular}
\end{table}

The cache primarily makes RL rollouts efficient and is optional at inference. Rendering oblique views on demand
increases the estimated total to 4.4 s/query, still about $7.3\times$ faster than CityAnchor's reported online
latency. The 2.3-hour figure covers oblique-cache construction only, not the complete deployment pipeline:
BEV generation, view-to-world calibration, and landmark registration are additional scene-preparation steps
whose individual construction times are not reported. These costs depend on the scene and are amortized over
queries; neither online ratio should be interpreted as an end-to-end deployment speedup.

\FloatBarrier
\endgroup

\section{Limitations}
\label{sec:limitations}

\method{} depends on a task-specific interactive environment: reliable \bev{} rendering and view-to-world calibration. Errors in calibration or rendering can project a plausible
2D box to the wrong world location, while absent, ambiguous, outdated, or missing landmarks force the agent to rely
more on visual search and local context, reducing efficiency.

The current policy submits a \bev{}-aligned bounding box, which is efficient and compatible with city-scale search but still
leaves room for improving fine-grained 3D segmentation accuracy. A natural extension is to expose multi-view 3D
instance segmentation models as additional tools, allowing the agent to refine the final prediction from a localized
2D region into a more accurate 3D instance mask or box.

\paragraph{Broader impacts and mitigation.}
Beyond beneficial applications such as navigation, urban planning, and accessibility, accurate language-driven
localization at city scale risks misuse for surveillance, persistent target tracking, sensitive-site identification,
or automated monitoring. We mitigate these risks by training only on public benchmarks (CityRefer, STPLS3D) without
personal identifiers, releasing code, checkpoints, and STPLS3D-Refer under a research-only license that prohibits
unconsented surveillance and individual tracking, and recommending privacy-preserving preprocessing (e.g., blurring
faces and license plates) for downstream adaptations.

\FloatBarrier

\endgroup

\endgroup

\end{document}